\documentclass[letterpaper]{article} 
\usepackage{aaai2026}  
\usepackage{times}  
\usepackage{helvet}  
\usepackage{courier}  
\usepackage[hyphens]{url}  
\usepackage{graphicx} 
\usepackage{natbib}  
\usepackage{caption} 
\usepackage{algorithm}
\usepackage{algorithmic}

\usepackage{newfloat}
\usepackage{listings}
\DeclareCaptionStyle{ruled}{labelfont=normalfont,labelsep=colon,strut=off} 
\floatstyle{ruled}
\newfloat{listing}{tb}{lst}{}
\floatname{listing}{Listing}
\usepackage{graphicx}
\usepackage{caption}  

\usepackage{booktabs}

\usepackage{array}
\usepackage{multirow} 

\usepackage{amsmath, amssymb}

\newcommand{\CLA}[1]{{\color[HTML]{4472c4} \textbf{#1}}}
\newcommand{\CLB}[1]{{\color[HTML]{E76254} \textbf{#1}}}

\usepackage{amssymb}
\usepackage{booktabs}
\newcommand{\xmark}{\ding{55}}  
\usepackage{pifont}            

\usepackage{pifont}
\newcommand{\cmark}{\textcolor{green!60!black}{\checkmark}}  
\usepackage{placeins}
\usepackage{tabularx}
\usepackage[table]{xcolor}

\usepackage{caption}

\usepackage{helvet}
\usepackage{courier}
\usepackage{pdfpages}

\usepackage{fancyhdr}
\usepackage{lipsum}

\fancypagestyle{secondpage}{
  \fancyhf{} 
  \fancyfoot[L]{\small Open-source dataset: \url{https://huggingface.co/datasets/xiaojiahao/StaticEmbodiedBench}}
}

\usepackage[absolute,overlay]{textpos}

\nocopyright

\title{Static and Plugged: Make Embodied Evaluation Simple}
\author{
    Jiahao Xiao\textsuperscript{\rm 1,2}\equalcontrib,
    Jianbo Zhang\textsuperscript{\rm 1,2}\equalcontrib,
    BoWen Yan\textsuperscript{\rm 1,3}\equalcontrib,
    Shengyu Guo\textsuperscript{\rm 1,4},
    Tongrui Ye\textsuperscript{\rm 1,5},
    Kaiwei Zhang\textsuperscript{\rm 1,2},\\
    Zicheng Zhang\textsuperscript{\rm 1,2},
    Xiaohong Liu\textsuperscript{\rm 2},
    Zhengxue Cheng\textsuperscript{\rm 2},
    Lei Fan\textsuperscript{\rm 2},
    Chuyi Li\textsuperscript{\rm 1,2}\thanks{Corresponding authors: Chunyi Li, Guangtao Zhai},
    Guangtao Zhai\textsuperscript{\rm 1,2}\footnotemark[2]
}
\affiliations{
    \textsuperscript{\rm 1}Shanghai AI Lab
    \textsuperscript{\rm 2}Shanghai JiaoTong University
    \textsuperscript{\rm 3}Tsinghua University
    \textsuperscript{\rm 4}Peking University
    \textsuperscript{\rm 5}Tongji University\\

    {Project Page: \url{https://staging.opencompass.org.cn/embodied-intelligence}}

}

\usepackage{bibentry}

\begin{document}
\maketitle

\begin{textblock}{4}(16.5,1)  
  \includegraphics[width=3cm]{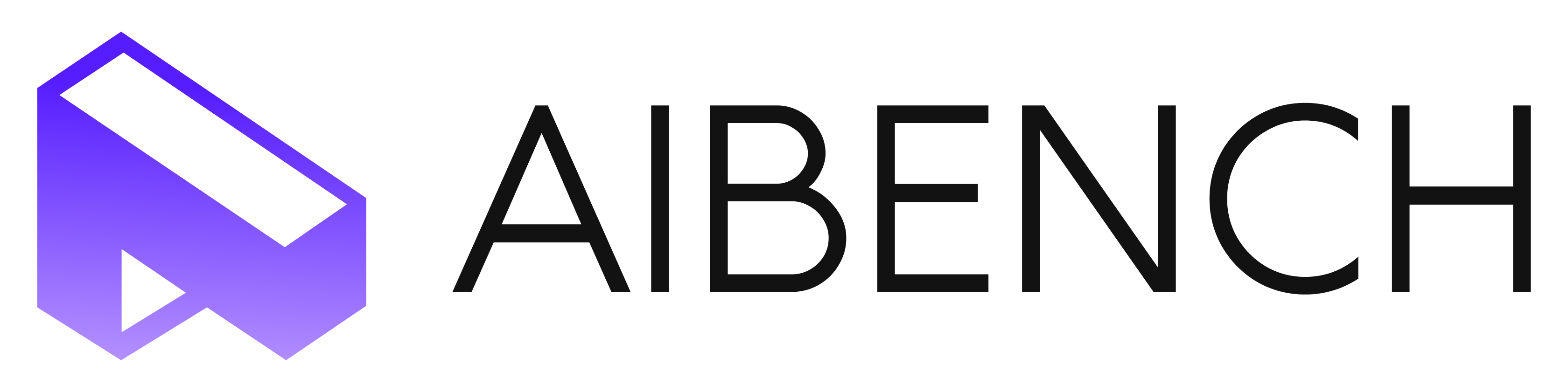}
\end{textblock}


\begin{abstract}
\begin{quote}

Embodied intelligence is advancing rapidly, driving the need for efficient evaluation. Current benchmarks typically rely on interactive simulated environments or real-world setups, which are costly, fragmented, and hard to scale. To address this, we introduce StaticEmbodiedBench, a plug-and-play benchmark that enables unified evaluation using static scene representations. Covering 42 diverse scenarios and 8 core dimensions, it supports scalable and comprehensive assessment through a simple interface. Furthermore, we evaluate 19 Vision-Language Models (VLMs) and 11 Vision-Language-Action models (VLAs), establishing the first unified static leaderboard for Embodied intelligence. Moreover, we release a subset of 200 samples from our benchmark to accelerate the development of embodied intelligence.

\end{quote}
\end{abstract}

\begin{figure*}[t] 
  \centering
  \includegraphics[width=\textwidth]{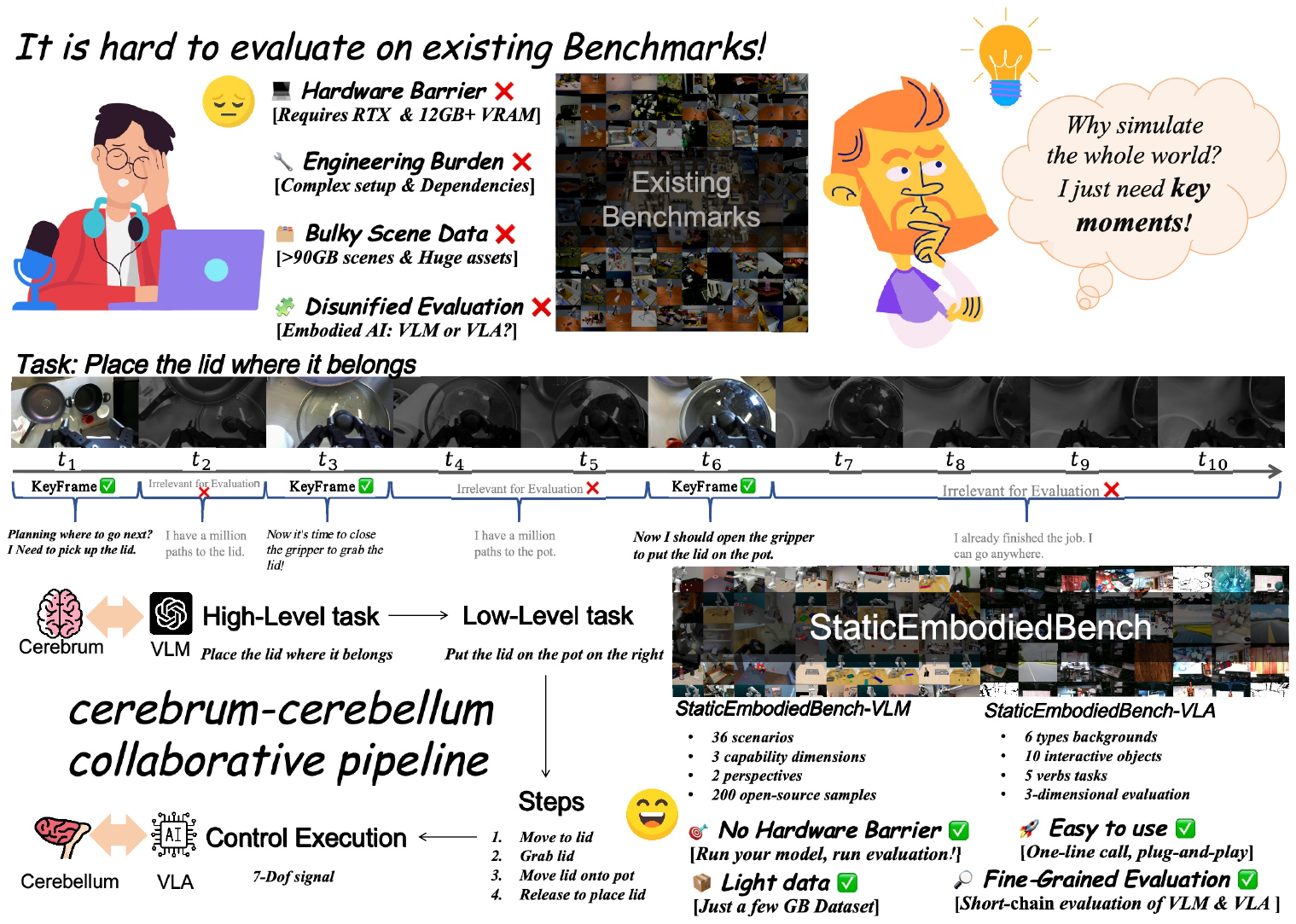}  
  \vspace{-7mm}
  \caption{Existing real-world and simulation-based embodied intelligence methods are hard to evaluate. By focusing on a few key moments, evaluation becomes much easier. We propose StaticEmbodiedBench — a simple yet comprehensive benchmark enabled by a cerebrum–cerebellum collaborative pipeline for assessing embodied intelligence.}
  \label{fig:spotlight}
\end{figure*}

\section{Introduction}
\label{sec:intro}

With the rapid development of Large Language Models (LLMs)~\cite{vlm:gpt_4o}, Vision-Language Models ~\cite{vlm:gemini-2.5,vlm:qwen2.5,vlm:internvl3}, and Vision-Language-Action models ~\cite{vla:octo,vla:pi0_v1,vla:rt_x,vla:openvla_v1}, embodied artificial intelligence (Embodied AI) is becoming increasingly capable of integrated perception, language understanding, and physical interaction. These systems are now widely applied into domains such as robotic manipulation~\cite{vla:cogact} and autonomous driving~\cite{vla:opendrivevla}. As a result, systematically evaluating embodied AI systems has become a central concern in the community.

Recently, researchers have proposed a number of benchmarks for evaluating embodied AI, such as ALFRED~\cite{dataset:alfred}, EmbodiedEval~\cite{dataset:embodiedeval}, and BEHAVIOR~\cite{dataset:behavior}. These benchmarks typically rely on highly realistic simulation platforms like LEGENT~\cite{platform:legent} and Habitat~\cite{platform:habitat,platform:habitat2,platform:habitat3} to recreate interactive tasks in realistic environments, forming closed-loop pipelines from observation input to action execution. Despite driving significant progress, these evaluation pipelines present two key limitations in practice, as shown in Figure~\ref{fig:spotlight}:

\textbf{Issue 1: Heavy reliance on simulation environments.} Many current benchmarks depend on massive scene datasets—often exceeding tens of gigabytes—and complex simulators, which pose significant barriers to use. For instance, Isaac Lab~\cite{dataset:isasslab}, one of the popular benchmarks, is based on Isaac Sim~\cite{platform:isaacsim}, which requires RTX GPUs with at least $8\,\mathrm{GB}$ of VRAM, involves a complicated installation process, and typically produces around $1\mathrm{GB}$ of data for accomplishing a single task. Such hardware requirements and engineering burdens significantly hinder the feasibility of conducting evaluations.

\textbf{Issue 2: Lack of unified evaluation.} Most benchmarks focus on only one type of embodied intelligence model. For example, EmbodiedEval~\cite{dataset:embodiedeval} evaluates the cognitive and decision-making abilities of VLMs, using image and language inputs to produce discrete actions. In contrast, RLBench~\cite{dataset:rlbench} targets the low-level control abilities of VLAs, outputting continuous 7-DoF signals. However, these benchmarks fail to evaluate the full pipeline that spans high-level reasoning and low-level execution, limiting insight into coordination and making it hard to identify which module limits overall performance.

To address these two challenges, we propose two key solutions: a novel static evaluation framework tailored for embodied AI to eliminate the need for simulation environments, and a cerebrum–cerebellum collaborative framework to unify the evaluation of both high-level reasoning and low-level control. For Issue 1, our key insight is that successful task execution often hinges on a small set of critical points within the agent’s trajectory. By identifying and isolating these keyframes, we can construct a lightweight, simulator-free benchmark that significantly reduces computational and engineering overhead, while retaining the ability to evaluate core embodied capabilities. For Issue 2, inspired by the “cerebrum–cerebellum collaboration” concept in cognitive science~\cite{li2025perceptualqualityassessmentembodied,vla:gr00tn1}, we decompose the embodied AI system into two components to evaluate: (1) Cerebrum component: implemented by VLMs, responsible for task comprehension, macro planning, and micro understanding based on vision-language input; (2) Cerebellum component: implemented by VLAs, responsible for executing actions and managing low-level control.

Building on these two insights, we introduce a unified static evaluation benchmark, StaticEmbodiedBench, as summarized in Table~\ref{tab:benchmark_comparison}, which separately assesses high-level \textit{Cognition} and \textit{Decision} abilities of VLMs and low-level \textit{Execution} capabilities of VLAs in embodied AI. Our framework relies solely on static images and textual metadata, enabling plug-and-play evaluation without the need for complex simulators, thereby significantly reducing deployment costs and technical barriers. Furthermore, we conduct comprehensive benchmarking across a broad range of state-of-the-art VLMs and VLAs, accompanied by in-depth analysis of the results. The main contributions are as follows:
\begin{itemize}
    \item We propose a novel static keyframe-based evaluation method and a cerebrum–cerebellum collaborative evaluation framework to decouple and assess VLM and VLA capabilities within embodied scenarios.
    \item Based on this, we build StaticEmbodiedBench, a \textbf{static}, \textbf{unified} and \textbf{plug-and-play} benchmark with 42 scenarios and 8 evaluation dimensions. We have partially released a subset of 200 samples from our benchmark.    
    \item We conduct extensive evaluations of 19 VLMs and 11 VLAs, establishing the first unified static leaderboard for embodied AI with thorough analysis.
\end{itemize}

\begin{table*}[t]
    \centering
    \renewcommand\arraystretch{1.0}
    \renewcommand\tabcolsep{7pt}
    \vspace{-6mm}
    \caption{
Comparison between StaticEmbodiedBench and existing benchmarks, covering whether data is \textbf{Static}, evaluation of \textbf{Cog}nition, \textbf{Dec}ision, and \textbf{Exe}cution, availability of \textbf{1st-view} and \textbf{3rd-view} perspectives, presence of \textbf{Sim}ulation and \textbf{Real}-world scenes, and the number of evaluation metrics for VLM and VLA indicated by \textbf{Dim}ension.
}
    \label{tab:benchmark_comparison}
    \resizebox{\textwidth}{!}{%
    \begin{tabular}{lcccccccccc}
        \toprule
        Benchmark & Venue & Static & Cog. & Dec. & Exe. & Dim. & 1st-view & 3rd-view & Sim. & Real. \\
        \midrule
        EQA~\cite{dataset:EQA}                   & CVPR 2018    & \xmark & \cmark & \cmark & \cmark & 1+1 & \cmark & \xmark & \cmark & \xmark \\
        ALFRED~\cite{dataset:alfred}             & CVPR 2020    & \xmark & \cmark & \cmark & \xmark & 6+0 & \cmark & \xmark & \cmark & \xmark \\
        ManiSkill~\cite{dataset:maniskill}       & ICLR 2023    & \xmark & \xmark & \xmark & \cmark & 0+1 & \cmark & \cmark & \cmark & \xmark \\
        BridgeData~\cite{dataset:bridgedata-v2}  & CoRL 2023    & \xmark & \xmark & \xmark & \cmark & 0+1 & \cmark & \cmark & \xmark & \cmark \\
        OpenEQA~\cite{dataset:openeqa}           & CVPR 2024    & \cmark & \cmark & \xmark & \xmark & 7+0 & \cmark & \xmark & \xmark & \cmark \\
        VLABench~\cite{dataset:vlabench}         & ArXiv 2024           & \xmark & \cmark & \cmark & \cmark & 6+1 & \cmark & \cmark & \cmark & \xmark \\
        EmbodiedEval~\cite{dataset:embodiedeval}& ArXiv 2025&\xmark &\cmark &\cmark &\xmark & 5+0&\cmark &\xmark & \cmark&\xmark\\
        \rowcolor{gray!15}
        StaticEmbodiedBench (Ours)              & --           & \cmark & \cmark & \cmark & \cmark & 5+3 & \cmark & \cmark & \cmark & \cmark \\
        \bottomrule
    \end{tabular}}
\end{table*}

\section{Method}
\thispagestyle{secondpage}
\subsection{Motivation: Static Keyframe-Based Evaluation}

Traditional evaluation of embodied intelligence typically relies on either real-world robotic deployment~\cite{vla:rt_x} or high-fidelity simulation environments~\cite{dataset:embodiedeval,dataset:behavior}. However, both approaches come with significant practical challenges. The former requires extensive human supervision and long-term testing, which is time-consuming and costly. The latter demands heavy computational resources, environment-specific configurations, and often special hardware setups (e.g., RTX GPUs required for simulators such as Isaac Sim~\cite{platform:isaacsim}), along with downloading large scene datasets. Interestingly, we observe that the success of most embodied tasks mainly depends on a small number of \textit{key points}, such as identifying the goal object or performing a grasp in the correct position. In contrast, how the agent navigates between these points—i.e., the trajectory—is often variable and not critical to task success (unless the task itself is related to navigation). Therefore, from an evaluation perspective, much of the information across the full interaction sequence is  redundant. Motivated by this insight, we propose a static, keyframe-based evaluation paradigm, which retains only keyframes corresponding to these critical states. Each image frame is paired with a customized question or task, and the model is evaluated solely based on its behavior at these checkpoint moments. This significantly reduces computational overhead while focusing on the core reasoning and control capabilities of embodied intelligence.

Additionally, to validate the effectiveness of static evaluation, it is important to measure its consistency with dynamic execution performance. Similar to the well-known Sim2Real gap in the simulation field~\cite{tobin2017domainrandomizationtransferringdeep}, we define a new metric, the \textbf{Static-to-Dynamic(S2D)} Gap, to capture the performance deviation between static keyframe-only evaluation and dynamic full-environment execution. Specifically, for a given model $m$ and task set $\mathcal{T}$, let $S_m(t)$ denote the model’s predicted score under static evaluation on task $t \in \mathcal{T}$, and $R_m(t)$ denote its real-world score based on the success rate conversion under dynamic execution. The S2D rate is defined as:
\[
\text{S2D}(m) = \mathrm{PLCC} \left( \{S_m(t)\}_{t \in \mathcal{T}}, \{R_m(t)\}_{t \in \mathcal{T}} \right),
\]
where $\mathrm{PLCC}(\cdot)$ denotes the Pearson Linear Correlation Coefficient between the static scores and the corresponding real-world scores. A higher S2D rate shows that keyframe-based evaluation better reflects the agent's true performance, thereby reflecting the benchmark's practical utility.

Moreover, many existing benchmarks tend to target disjoint components of the embodied system—some focus solely on VLMs~\cite{dataset:embodiedeval} as high-level planners, while others evaluate VLAs~\cite{dataset:maniskill} as low-level executors, without considering their interplay, which limits a holistic understanding and evaluation of embodied intelligence. To address this gap, we argue for a unified evaluation framework that spans the entire decision-to-action pipeline, from abstract task planning to grounded motor execution. Motivated by this need, we conceptualize embodied systems through a \textit{cerebrum–cerebellum collaborative pipeline} view (see next section for details), which disentangles cognitive planning and motor execution as two distinct yet interdependent stages. This perspective guides the design of two targeted evaluation components within our benchmark: StaticEmbodiedBench-VLM and StaticEmbodiedBench-VLA as shown in Figure~\ref{fig:spotlight}.


\begin{figure*}[!t] 
  \centering
  \includegraphics[width=\textwidth]{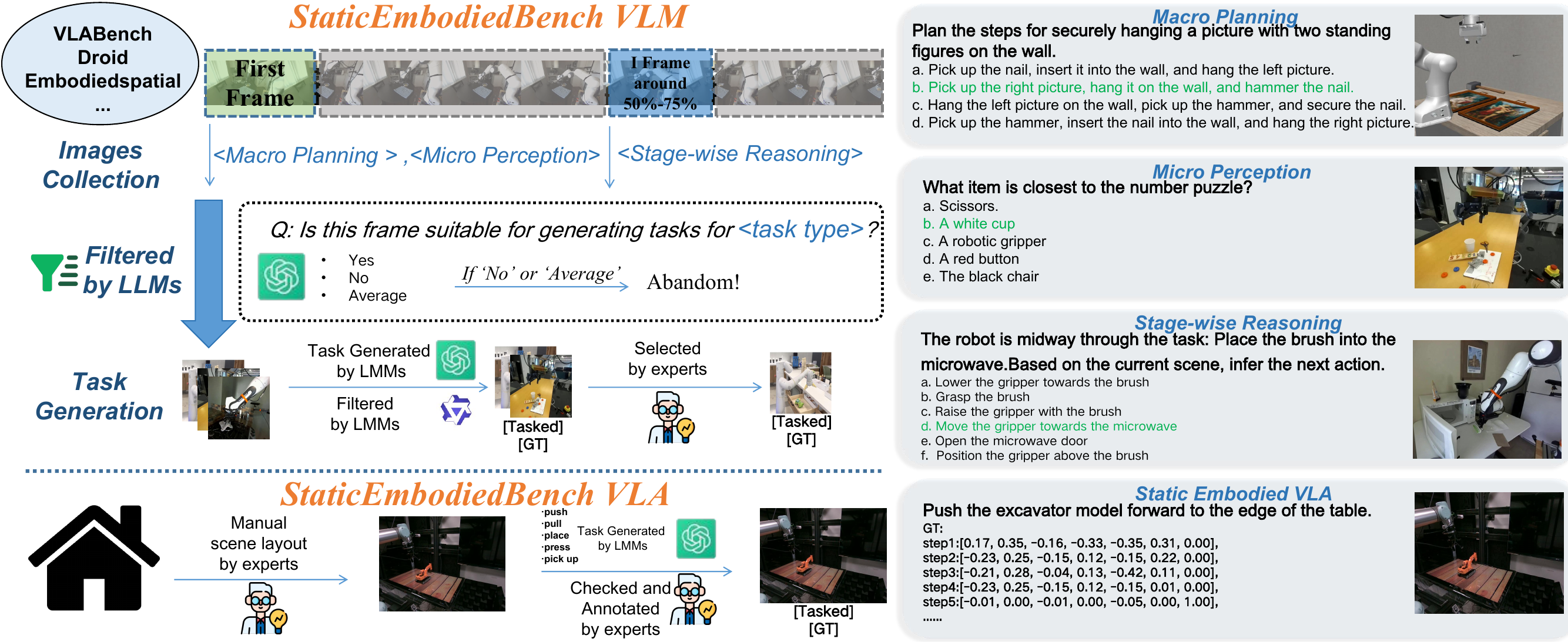}  
  \caption{Overview of the StaticEmbodiedBench dataset construction pipeline (left) and representative examples (right). }
  \label{fig:method}
\end{figure*}
\subsection{Design: Cerebrum–Cerebellum Evaluation}
\label{sec:design_cerebrum_cerebellum}
Inspired by recent advances in embodied cognition pipelines~\cite{li2025perceptualqualityassessmentembodied,vla:gr00tn1}, we adopt a cerebrum–cerebellum division of labor to model how embodied agents solve complex tasks. Specifically, a Vision-Language Model first interprets a high-level instruction and plans a sequence of interpretable sub-tasks—serving as the ``cerebrum'' of the system. Then, a Vision-Language-Action Model executes these sub-tasks step-by-step, acting as the ``cerebellum''. This two-stage pipeline achieves superior task success by leveraging the respective strengths of abstract reasoning and low-level control~\cite{vla:saycan}. Motivated by this, we propose evaluating each module independently by designing two short-chain benchmarks that assess their functional competence separately, rather than evaluating the full end-to-end system jointly. This separation enhances interpretability and diagnostic insight, and aligns with how capabilities are developed and improved in modular systems.

\subsubsection{StaticEmbodiedBench-VLM Design}

To evaluate the role of Vision-Language Models as the cognitive core of embodied intelligence, we design StaticEmbodiedBench-VLM to probe three core cognitive dimensions, each tested under two different visual perspectives:

\begin{itemize}
    \item\textbf{Macro Planning:} Can VLMs break down a high-level instruction into an interpretable sequence of sub-tasks? This evaluates their ability to perform structured long-horizon planning before execution.
    
    \item \textbf{Micro Perception:} Can VLMs recognize fine-grained visual cues such as spatial arrangements, object types, and affordances? This tests perception capabilities necessary for correct sub-task grounding.
    
    \item \textbf{Stage-wise Reasoning:} Can VLMs identify the current step in the sub-task sequence and determine the next optimal action? This skill is essential for dynamic task tracking and mid-execution adaptation.
    
    \item \textbf{First-Person View:}  Observations captured from the end-effector mounted egocentric perspective, testing the model's abilities under egocentric context.
    
    \item \textbf{Third-Person View:} External or top-down observations provide a global view of the environment, testing the model's abilities under exocentric context.
\end{itemize}

\subsubsection{StaticEmbodiedBench-VLA Design}

To evaluate execution capabilities of the VLA module, we model each motion command as a 7-DoF action vector:
\[
\hat{\mathbf{x}} = (x, y, z, \alpha, \beta, \gamma, s),
\]
where $(x, y, z)$ is the Cartesian position, $(\alpha, \beta, \gamma)$ are orientation angles (Euler representation), and $s$ denotes the gripper state. Each predicted motion step $\hat{\mathbf{x}}$ is compared with the expert reference trajectory $\mathbf{x}^{\ast}$ using the L2 loss:
\[
\mathcal{L}_{\text{exec}} = \left\| \hat{\mathbf{x}} - \mathbf{x}^{\ast} \right\|_2,
\]
and to enhance interpretability, we decompose the error into:
\begin{align*}
\mathrm{e}_{\mathrm{position}} &= \left\| (x, y, z) - (x^\ast, y^\ast, z^\ast) \right\|_2, \\
\mathrm{e}_{\mathrm{orientation}} &= \left\| (\alpha, \beta, \gamma) - (\alpha^\ast, \beta^\ast, \gamma^\ast) \right\|_2, \\
\mathrm{e}_{\mathrm{end\text{-}effector}} &= |s - s^\ast|.
\end{align*}
This fine-grained decomposition provides deeper insight into execution errors—such as failure to reach target positions or misaligned grasp orientation—enabling more precise evaluation of the physical motor competence of VLAs.

\subsection{Benchmark Construction}
\subsubsection{StaticEmbodiedBench-VLM}
To build a high-quality benchmark evaluating the cognitive abilities of VLMs across diverse embodied contexts, we design a multi-stage pipeline involving large-scale keyframe sampling, automatic filtering, task generation, and human validation (see Figure~\ref{fig:method}).

We begin by collecting over 300,000 high-resolution frames from 36 scenarios spanning five high-quality datasets: \textit{Droid}~\cite{dataset:droid}, \textit{VLABench}~\cite{dataset:vlabench}, \textit{EmbSpatial-Bench}~\cite{dataset:embodiedspatial}, \textit{EmbodiedBench}~\cite{dataset:embodiedbench}, and \textit{SAT}~\cite{dataset:sat}. These frames capture rich visual semantics across diverse scenes, objects, and tasks. Furthermore, to target three specific capabilities, we employ sampling strategies tailored to each:
\begin{itemize}
  \item For Macro Planning and Micro Perception tasks, we select the first step frame from each task sequence, which reflects VLM’s initial perception and planning context. This keyframe provides the model with high-level task information and the initial environment state.

  \item For Stage-wise Reasoning tasks, we sample frames occurring approximately between 50\% to 75\% of the task timeline, representing intermediate task states where mid-task reasoning is crucial, and only I-frames are extracted in this process to ensure high visual fidelity.
\end{itemize}

Each image is passed through a GPT-based classifier that predicts its suitability for Macro, Micro, or Stage reasoning tasks. We assign one of three tags: \textit{Yes}, \textit{No}, or \textit{Average}. Only images labeled \textit{Yes} are retained, narrowing the candidate set from 300,000 to around 4,000 images with strong task relevance and clarity. Next, we design expert-crafted prompts and feed the 4,000 filtered images into GPT-4o, guiding it to generate high-quality question–answer pairs aligned with each capability dimension. The prompts are customized for each dimension, ensuring the generated tasks are diverse and meaningful. Last but not least, all generated tasks undergo a final review process by human annotators. Tasks are corrected, refined, or discarded based on the quality of the labeled tasks. After this multi-stage validation, we curate a final set of 1,000 high-quality task samples that form the StaticEmbodiedBench-VLM benchmark.

\subsubsection{StaticEmbodiedBench-VLA}
\begin{table*}[!t]
    \centering
    \renewcommand\arraystretch{1.0}
    \renewcommand\tabcolsep{3pt}
    \caption{Evaluation results of 19 VLMs and 11 VLAs on StaticEmbodiedBench. For VLMs, Dim-1, Dim-2, and Dim-3 correspond to Macro Planning, Micro Perception, and Stage-wise Reasoning, respectively. For VLAs, Dim-1, Dim-2, and Dim-3 correspond to Position, Orientation, and End-effector control.The columns First, Third, and Score represent the model's performance under First-Person View, Third-Person View, and the Overall Score. [Keys: \CLA{Best}/\CLB{Second best}/\colorbox{gray!20}{Worst} in group]}
    \label{tab:result}
    \resizebox{\linewidth}{!}{
\begin{tabular}{l|l|cc|c|ccccc|c}
\toprule
Group & Model & Params & LLM  & Steps & Dim-1$\uparrow$& Dim-2$\uparrow$ & Dim-3$\uparrow$ & First$\uparrow$ & Third$\uparrow$ & Score$\uparrow$ \\
\midrule
\multicolumn{11}{c}{\textit{Vision-Language Models result}} \\

\midrule
\multirow{10}{*}{Closed}
& ChatGPT-4o-latest~\cite{vlm:gpt_4o}       & N/A   & --           & --                 & \CLB{87.33} & \CLB{74.00} & \CLA{52.86} & \CLB{59.90} & \CLA{62.01} & \CLA{61.20} \\
& GPT-4.1~\cite{vlm:gpt-4.1}                 & N/A   & --           & --                 & \CLA{88.00} & 70.00 & \CLB{52.60} & \CLA{60.20} & \CLB{60.70} & \CLB{60.50} \\
& Gemini-1.5-pro~\cite{vlm:gemini-1.5}          & N/A   & --           & --                & 82.67 & 72.67 & 51.29 & 57.79 & 60.06 & 59.20 \\
& Claude-sonnet-4 ~\cite{vlm:claude-4}        & N/A   & --           & --                & 86.00 & 70.67 & 49.14 & 58.60 & 57.55 & 57.90 \\
& GPT-4.1-mini~\cite{vlm:gpt-4.1}            & N/A   & --           & --               & 78.00 & 65.33 & 51.00 & 57.01 & 57.29 & 57.20 \\
& GPT-4o~\cite{vlm:gpt_4o}                  & N/A   & --           & --               & 85.33 & 74.00 & 44.57 & 54.22 & 55.73 & 55.10 \\
& Gemini-2.5-preview~\cite{vlm:gemini-2.5}      & N/A   & --           & --               & 78.67 & \CLA{76.67} & 45.29 & 50.81 & 57.63 & 55.00 \\
& GPT-4o-mini~\cite{vlm:gpt_4o}             & N/A   & --           & --                & 74.00 & 58.00 & 48.71 & 52.76 & 55.73 & 53.90 \\
& Grok-2-vision~\cite{vlm:grok-2}           & N/A   & --           & --               & 79.33 & 54.67 & 45.57 & 49.84 & 55.47 & 52.00 \\
& GPT-4.1-nano~\cite{vlm:gpt-4.1}            & N/A   & --           & --               & \colorbox{gray!20}{73.33} & \colorbox{gray!20}{37.33} & \colorbox{gray!20}{43.43} & \colorbox{gray!20}{47.08} & \colorbox{gray!20}{46.88} & \colorbox{gray!20}{47.00} \\
\cmidrule{1-11}
\multirow{9}{*}{Open}
& InternVL2.5-78B-MPO~\cite{vlm:internvl2.5}     & 78B   & Qwen2.5-72B   & -- & \CLA{88.67} & \CLB{80.00} & \CLA{53.14} & \CLA{61.69} & \CLA{62.99} & \CLA{62.50} \\
& InternVL3-78B~\cite{vlm:internvl3}           & 78.4B & Qwen2.5-72B   & -- & \CLB{88.67} & 80.00 & 51.86 & \CLB{61.69} & 61.53 & \CLB{61.60} \\
& Qwen2.5-VL-72B~\cite{vlm:qwen2.5}          & 73.4B & Qwen2.5-72B           & -- & 87.33 & 73.33 & \CLB{52.43} & 60.94& 60.71  & 60.80 \\
& InternVL3-38B~\cite{vlm:internvl3}           & 38.4B & Qwen2.5-32B  & -- & 85.33 & 78.67 & 51.29 & 59.42 & 61.20 & 60.50 \\
& InternVL2.5-38B-MPO~\cite{vlm:internvl2.5}     & 38B   & Qwen2.5-32B   & -- & 82.00 & \CLA{82.00} & 51.14 & 57.79 & \CLB{62.01} & 60.40 \\
& Qwen2.5-VL-32B~\cite{vlm:qwen2.5}          & 33.5B & Qwen2.5-32B           & -- & 84.00 & 74.00 & 48.71 & 55.19 & 59.42 & 57.80 \\
& Llama-4-Scout~\cite{vlm:Llama-4-Scout}           & 109B  & --                          & -- & 78.67 & 50.67 & 45.86 & 52.60& 50.81  & 51.50 \\
& LLava-v1.5-7B~\cite{vlm:llava2}           & 7.2B  & Vicuna-1.5-7B    & -- & 56.00 & \colorbox{gray!20}{37.33} & 42.00 & 46.92 & 41.23 & 43.40 \\
& InternVL3-1B~\cite{vlm:internvl3}            & 0.94B & Qwen2.5-0.5B    & -- & \colorbox{gray!20}{53.33} & 42.00 & \colorbox{gray!20}{40.00} & \colorbox{gray!20}{46.10} & \colorbox{gray!20}{39.94} & \colorbox{gray!20}{42.30} \\
\midrule
\multicolumn{11}{c}{\textit{Vision-Language-Action Models Result}} \\

\midrule
\multirow{11}{*}{Open}
& Octo-base-1.5~\cite{vla:octo}& 93M   & --                     & 4   & \CLA{70.37} &\CLB{49.17} & \CLB{49.17 } & --   & --   & \CLA{58.25} \\
& Octo-small-1.5~\cite{vla:octo}            & 27M   & --                      & 4   & \CLB{67.22} & 43.76  & 4.000  & --   & --   & \CLB{48.14} \\
& OpenVLA-7B ~\cite{vla:openvla_v1}                     & 7B    & Llama-2-7B             & 1   & 59.13  & 44.27  & 16.85 & --   & --   & 46.72 \\
& Pi0-fast-droid ~\cite{vla:pi0_fast}               & 470M  & DistilBERT                   & 10  & 22.29  & 33.12  & \CLA{74.15} & --   & --   & 34.34 \\
& OpenVLA-7B-oft-libero-spatial~\cite{vla:openvla_v2}  & 7B    & Llama-2-7B            & 1   & 7.851  & \CLA{53.11} & 24.19  & --   & --   & 29.58 \\
& CogACT-Large~\cite{vla:cogact}                   & 7.3B  & Prismatic-7B       & 16  & 29.59  & 23.53  & \colorbox{gray!20}{0.492}  & --   & --   & 22.84 \\
& CogACT-Base ~\cite{vla:cogact}                     & 7B    & Prismatic-7B            & 16  & 30.56  & 21.72 & 0.634   & --   & --   & 22.50 \\
& Openvla-7B-oft-libero-object~\cite{vla:openvla_v3}   & 7B    & Llama-2-7B             & 1   & 7.180   & 37.09  & 15.03 & --   & --   & 21.12  \\
& Pi0-droid~\cite{vla:pi0_v1}                    & 3.3B  & PaliGemma-3B            & 10  & 17.39  & 22.53  & 18.72 & --   & --   & 19.78  \\
& Pi0-libero-fast ~\cite{vla:pi0_v2}           & 470M  & DistilBERT                   & 50  & \colorbox{gray!20}{5.742}  & 14.11  & 0.811   & --   & --   & 8.620   \\
& Pi0-libero~\cite{vla:pi0_v3}                  & 3.3B  & PaliGemma-3B           & 50  & 11.49  & \colorbox{gray!20}{3.333 }  & \colorbox{gray!20}{0.533 }  & --   & --   & \colorbox{gray!20}{6.431 }  \\
\bottomrule
\end{tabular}
}
\end{table*}

We construct the dataset entirely through real-world robotic demonstrations with background images rendered directly on the tabletop display. Specifically, we design 100 tabletop manipulation tasks and execute them with a UR5 robotic arm in a controlled lab setting. Each task is composed by combining:
\begin{itemize}
    \item 6 background contexts: \textit{cluttered desktop}, \textit{varied table textures}, \textit{road surface}, \textit{underwater scene}, \textit{nuclear station}, \textit{indoor environment}.
    \item 10 diverse objects: \textit{cup}, \textit{doll-1}, \textit{doll-2}, \textit{car-model-1}, \textit{car-model-2}, \textit{LEGO-1}, \textit{LEGO-2}, \textit{submarine model}, \textit{workpiece}, \textit{excavator model}.
    \item 5 interaction verbs: \textit{push}, \textit{pull}, \textit{place}, \textit{pick up}, \textit{press}.
\end{itemize}

A human expert directly tele-operates the UR5 arm to accomplish each task. We record the full trajectory of the 7-DoF motion $a_t$ at 50 evenly sampled steps per task, and also record the first-person, third-person, and depth images of the entire process as $o_t$ through multiple cameras. Each motion step is saved as a frame in the sequence:
\[
(\mathbf{o}_1, \mathbf{a}_1), (\mathbf{o}_2, \mathbf{a}_2), \dots, (\mathbf{o}_{50}, \mathbf{a}_{50}).
\]
The result is a dataset of 100 static images, each paired with an initial task instruction and its corresponding expert tra-jectory. This benchmark provides a high-quality reference for evaluating the "cerebellum" capabilities of VLA models in pre-cise and diverse manipulation contexts.

\begin{figure*}[t]
    \centering
    \includegraphics[width=\textwidth]{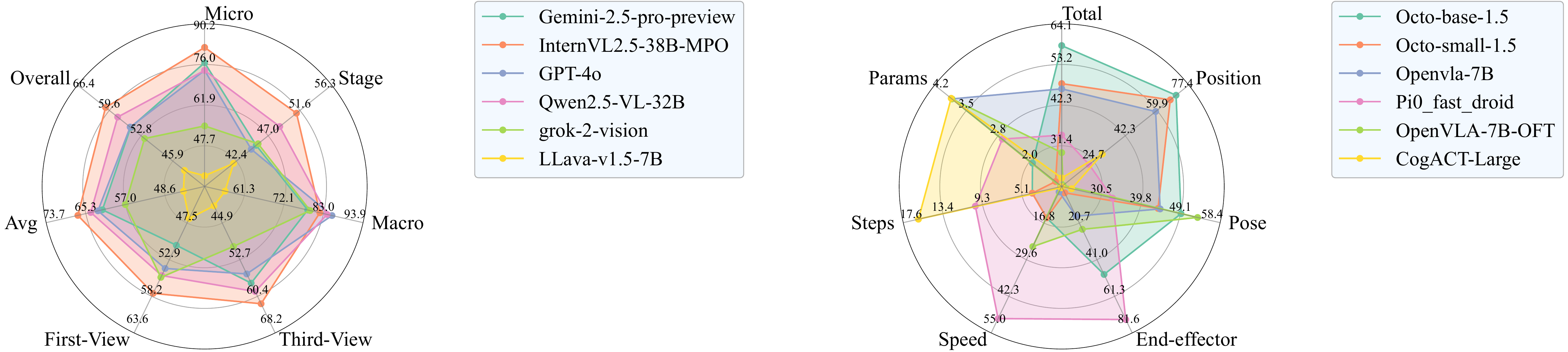}
    \vspace{-4mm}
    \caption{Radar charts showing performance of VLMs (left) and VLAs (right) on StaticEmbodiedBench. (Zoom in for detail)}
    \label{fig:radar_chart}
\end{figure*}

\vspace{-4mm}

\begin{figure*}[t]
    \centering
    \includegraphics[width=\textwidth]{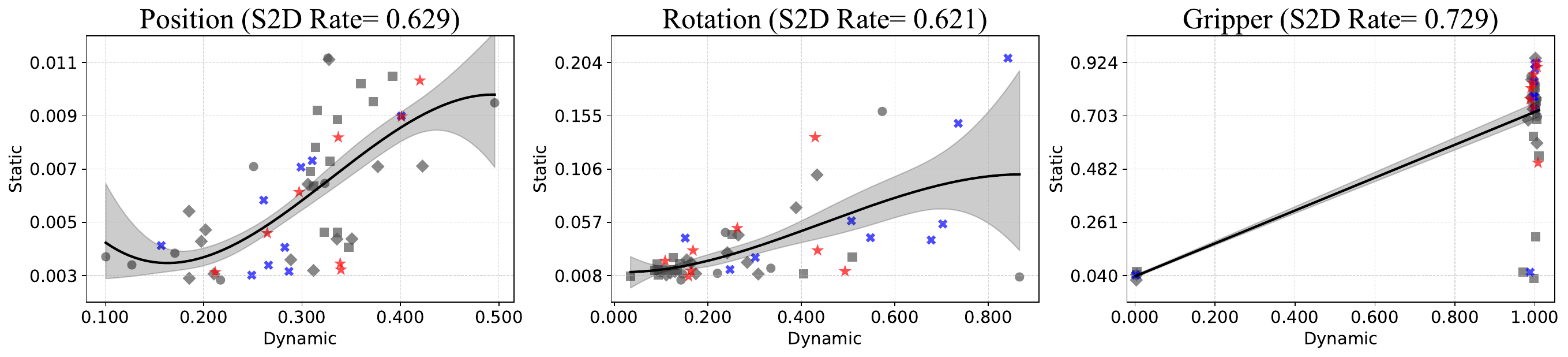}
    \vspace{-7mm}
    \caption{Scatter plots of the Octo model’s S2D rate on 50 samples for Position, Rotation, and Gripper. Marker shapes indicate task types: with \textit{Pick up} as circle, \textit{Place} as square, \textit{Press} as diamond, \textit{Pull} as X, and \textit{Push} as star. Curves show fitted trends. The gray region represents the ±1 standard deviation interval of the bootstrap-fitted curves.}
    \label{fig:plcc}
\end{figure*}

\section{Experiment and result}

In this part, we conduct large-scale evaluations of state-of-the-art models using StaticEmbodiedBench(see Table~\ref{tab:result} and radar chart in Figure~\ref{fig:radar_chart}), accompanied by in-depth analysis. In addition, we assess the effectiveness of our static benchmark through measuring its S2D rate.

\subsection{VLM Evaluation}

For the StaticEmbodiedBench-VLM, we integrated it into the VLMEvalKit~\cite{platform:vlmevalkit}, which not only enables one-line code to easily complete the evaluation, but also supports the circular evaluation to improve the accuracy~\cite{dataset:mmbench}. In the experiment, with two NVIDIA A800 GPUs, we evaluated 19 popular models - 10 closed-source models including ChatGPT-4o-latest, GPT-4.1, GPT-4.1-mini, GPT-4.1-nano, GPT-4o and GPT-4o-mini~\cite{vlm:gpt_4o, vlm:gpt-4.1}, Gemini-1.5-pro~\cite{vlm:gemini-1.5}, Gemini-2.5-preview~\cite{vlm:gemini-2.5}, Claude-sonnet-4~\cite{vlm:claude-4}, and Grok-2-Vision~\cite{vlm:grok-2}, and 9 representative open-source models, including InternVL2.5-78B-MPO and InternVL2.5-38B-MPO~\cite{vlm:internvl2.5}, InternVL3-78B, InternVL3-38B, and InternVL3-1B~\cite{vlm:internvl3}, Qwen2.5-VL-72B and Qwen2.5-VL-32B~\cite{vlm:qwen2.5}, LLaVA-v1.5-7B~\cite{vlm:llava2}, and LLaMA-4-Scout~\cite{vlm:Llama-4-Scout}.

In our evaluation, InternVL2.5-78B-MPO achieved the highest overall score of 62, followed by InternVL3-78B and GPT-4o. Notably, InternVL2.5-78B-MPO excelled across multiple dimensions, including macro planning (88.7), stage reasoning (53.1), and achieved top scores in both third-person (63.0) and first-person (61.7) visual perspectives. Meanwhile, InternVL2.5-38B-MPO demonstrated outstanding micro perception performance (82.0), highlighting the promise of compact models for detailed spatial reasoning.

Interestingly, despite sharing the same model size (78B), InternVL2.5-MPO consistently outperformed InternVL3, suggesting that the MPO (Mixed Preference Optimization)~\cite{wang2025enhancingreasoningabilitymultimodal} strategy is particularly effective in embodied settings. Moreover, a clear trend emerges across the three evaluation dimensions: macro planning tasks are significantly easier for current VLMs, whereas stage-wise reasoning remains the most challenging. This suggests that models excel at initial goal inference and global planning—likely benefiting from large-scale pretraining on general world knowledge—but struggle with mid-execution reasoning, where they must adapt to dynamic environmental changes and determine the next specific action based on the current state. This highlights a fundamental limitation in VLMs’ ability to handle real-time, situated cognition and decision-making tasks in embodied contexts.

\subsection{VLA Evaluation}
For the StaticEmbodiedBench-VLA, we evaluate VLAs by measuring the L2 distance between the model-generated 7-DoF control trajectories and expert-labeled ground truth, recorded through teleoperation of a UR5 robotic arm. Scoring is designed such that an L2 distance \(\leq\)1mm earns a full 100 points, while a distance \(\geq\)1m receives 0, with intermediate distances mapped logarithmically between 0 and 100. On a single A800 GPU, we evaluated 11 open-source VLA models, including Octo-base-1.5 and Octo-small-1.5~\cite{vla:octo}, OpenVLA-7B~\cite{vla:openvla_v1}, OpenVLA-7B-OFT-Libero-Spatial and OpenVLA-7B-OFT-Libero-Object~\cite{vla:openvla_v2, vla:openvla_v3}, pi0-fast-droid~\cite{vla:pi0_fast}, pi0-droid~\cite{vla:pi0_v1}, pi0-libero-fast and pi0-libero~\cite{vla:pi0_v2, vla:pi0_v3}, and CogACT-Base and CogACT-Large ~\cite{vla:cogact}.

In our evaluation, Octo-base-1.5 topped the leaderboard with a total score of 58.3, outscoring the runner-up by over 10 points. Dimension-wise, Octo-base-1.5 led in position accuracy (70.4), OpenVLA-7B-OFT-Libero-Spatial excelled in orientation accuracy (53.1), and Pi0-fast-droid ranked highest in gripper accuracy (74.2), demonstrating differentiated strengths across manipulation subtasks. Interestingly, we observed that VLA models incorporating large language models, such as OpenVLA and CogACT, did not outperform lighter models using standard Transformers. We hypothesize that this is due to our short-chain, cerebellum-only evaluation setup, which isolates the VLA component from complex reasoning. Since the inputs to VLAs in our benchmark are already easy and specific sub-task, the advantages of powerful LLMs may not be fully activated. As a result, smaller and faster models can perform competitively, or even better, under this decoupled evaluation framework.

\subsection{Cost Comparison}
In our experiments, we conducted a systematic comparison of static evaluation, simulation-based evaluation (e.g., Isaac Lab), and real-world evaluation (e.g., UR5), across four aspects: hardware cost, data volume per task, preparation time, and per-sample evaluation time, as shown in Table~\ref{tab:evaluation_comparison}. The results show that static evaluation requires significantly less investment in terms of money, manpower, and time, greatly lowering the threshold for engaging with this field and making the evaluation more accessible. In addition, the static evaluation method enables \textit{“Run your model, run evaluation”}, without the need for special hardware support. In comparison, Isaac Lab simulation requires high-performance GPUs such as the RTX 4090, and real-world evaluation demands a full UR5 robotic arm, both of which bring high hardware costs. In terms of data, simulation for a single scenario often results in several gigabytes of data, and real-world evaluation requires on-site environment setup. In contrast, static evaluation only requires kilobyte-level data, which is extremely lightweight, reducing storage pressure and enabling more comprehensive and diverse evaluations under the same data budget. Regarding preparation, simulation environments are complex and may require hours to days of debugging, while real-world systems often take weeks to set up. Static evaluation, however, can be prepared within minutes, greatly reducing the burden on engineers. In terms of evaluation efficiency, static evaluation only takes about 0.1 seconds per sample, much faster than the 10+ seconds per task required for simulation or real-world evaluation. This makes static evaluation the preferred, or even the only feasible, option in scenarios requiring the evaluation of a large number of models or tasks.

\subsection{Static-to-Dynamic Gap Validation}

In this part, we validate the effectiveness of our static evaluation framework, as shown in Figure~\ref{fig:plcc}. In particular, for high-level Cognition and Decision tasks, which are inherently discrete in nature, the static-to-dynamic gap can often be negligible. For low-level Execution tasks, however, we further measure the S2D rate—the correlation between static evaluation scores and real-world interactive scores—to quantitatively assess the reliability of our static dataset. We conduct an S2D evaluation on the StaticEmbodiedBench-VLA benchmark using the Octo ~\cite{vla:octo}model. Specifically, we uniformly sample 50 representative task instances across different task types and deploy them on a real UR5 robot. For each instance, full end-to-end physical execution is performed, and human annotators record the real-world success score for each interaction.

Meanwhile, we compute the static evaluation scores for the same 50 instances using our frame-wise prediction and trajectory comparison method. The correlation between static and real-world scores is then quantified using the PLCC metric. As a result, the S2D is reported across three control dimensions: 0.629 for position, 0.621 for rotation, and 0.729 for end-effector. The average S2D across all dimensions is 0.66, indicating a reasonably strong linear relationship between static evaluation and actual execution performance. A higher S2D rate suggests that the static metric is a reliable proxy for dynamic performance. With an average S2D of 0.66, our results support the feasibility and validity of using static, keyframe-based evaluation for embodied action models in real-world settings.

\section{Discussion}
We discuss three key aspects that highlight the strengths of our approach and outline directions for future development. 

Our benchmark showcases diverse scenes and tasks in tabletop manipulation, providing strong evidence that static evaluation can capture complex embodied capabilities. This lays the groundwork for extending to broader domains—like navigation, locomotion, and autonomous driving—where static representations remain underexplored. Moreover, while VLM evaluation has benefited from standardized toolkits such as VLMEvalKit, VLA evaluation remains fragmented due to the lack of an integrated suite. Our plug-and-play evaluation paradigm, based entirely on static data, enables flexible and simulator-free assessment of VLA models. We believe this simplicity can serve as the foundation for a unified and extensible evaluation toolkit—VLAEvalKit—that standardizes and streamlines VLA evaluation across diverse tasks and embodiments.

Another key challenge lies in aligning static evaluation with dynamic execution. While our current benchmark provides a promising baseline, further gains could be achieved by automatically identifying task-specific keyframes—such as those corresponding to goal localization, contact initiation, or task completion—that capture the most informative visual cues. This presents an exciting opportunity for future work on adaptive keyframe selection and S2D alignment.

\begin{table}[!t]
\centering
\caption{Comparison of Embodied AI evaluation paradigms regarding hardware cost, data size and efficiency (per task).}
\label{tab:evaluation_comparison}
\renewcommand\arraystretch{1.0}
\renewcommand\tabcolsep{6.5pt}
\begin{tabular}{lccc}
\toprule
\textbf{Aspect} & \textbf{Static} & \textbf{Simulation} & \textbf{Real-Robot} \\
\midrule
Hardware & \$0 & \$100--\$1,000 & \$10,000+ \\
Data Size & \textless 1Mb &1-5~GB & On-site \\
Setup Time & Minutes & Hours--Days & Days--Weeks \\
Eval Time & $\sim$0.1~s & $\sim$10~s & $\sim$10~s \\
\bottomrule
\end{tabular}
\end{table}

\section{Conclusion}

In this paper, we propose a cerebrum–cerebellum collaborative evaluation framework along with the StaticEmbodiedBench dataset, which evaluates both high-level Cognition and Decision abilities in VLMs and low-level Execution capabilities in VLAs. By systematically evaluates 19 state-of-the-art VLMs and 11 VLAs, we publish the first unified static leaderboard for Embodied intelligence. Additionally, we perform real-robot experiments to measure the Static-to-Dynamic gap of StaticEmbodiedBench, validating the feasibility of our static evaluation approach. In addition, we release a subset of 200 annotated examples to facilitate open research and reproducibility. Overall, StaticEmbodiedBench provides a low-cost, simple, and systematic way to evaluate embodied intelligence, and we sincerely hope our work will accelerate the development of embodied intelligence.

\section*{Acknowledgements}

This work is produced by the Evaluation Group at Shanghai AI Laboratory, aiming to standardize evaluation protocols for artificial intelligence. We thank all contributors to the project for their support and insights. For more information, please refer to the AI-Bench project~\cite{aibench}.

\bibliography{aaai2026}

\clearpage


\section*{Supplementary material}

# Appendix

## 1 Scope and Future Work

Our benchmark focuses on tabletop manipulation tasks and is validated within this scope. Its use in other embodied AI areas like navigation or autonomous driving needs further study and adaptation. To encourage future work and collaboration, we provide a clear direction and demo for static evaluation of embodied AI tasks. We hope this serves as a foundation inspiring the community to expand dataset scale, task diversity, and enhance automated keyframe detection and alignment between static evaluation and dynamic execution.

## 2 Case study

We showcases example cases for each evaluation dimension in our benchmark in Figure 1. More example cases can be found in our open-source dataset containing 200 evaluation questions.

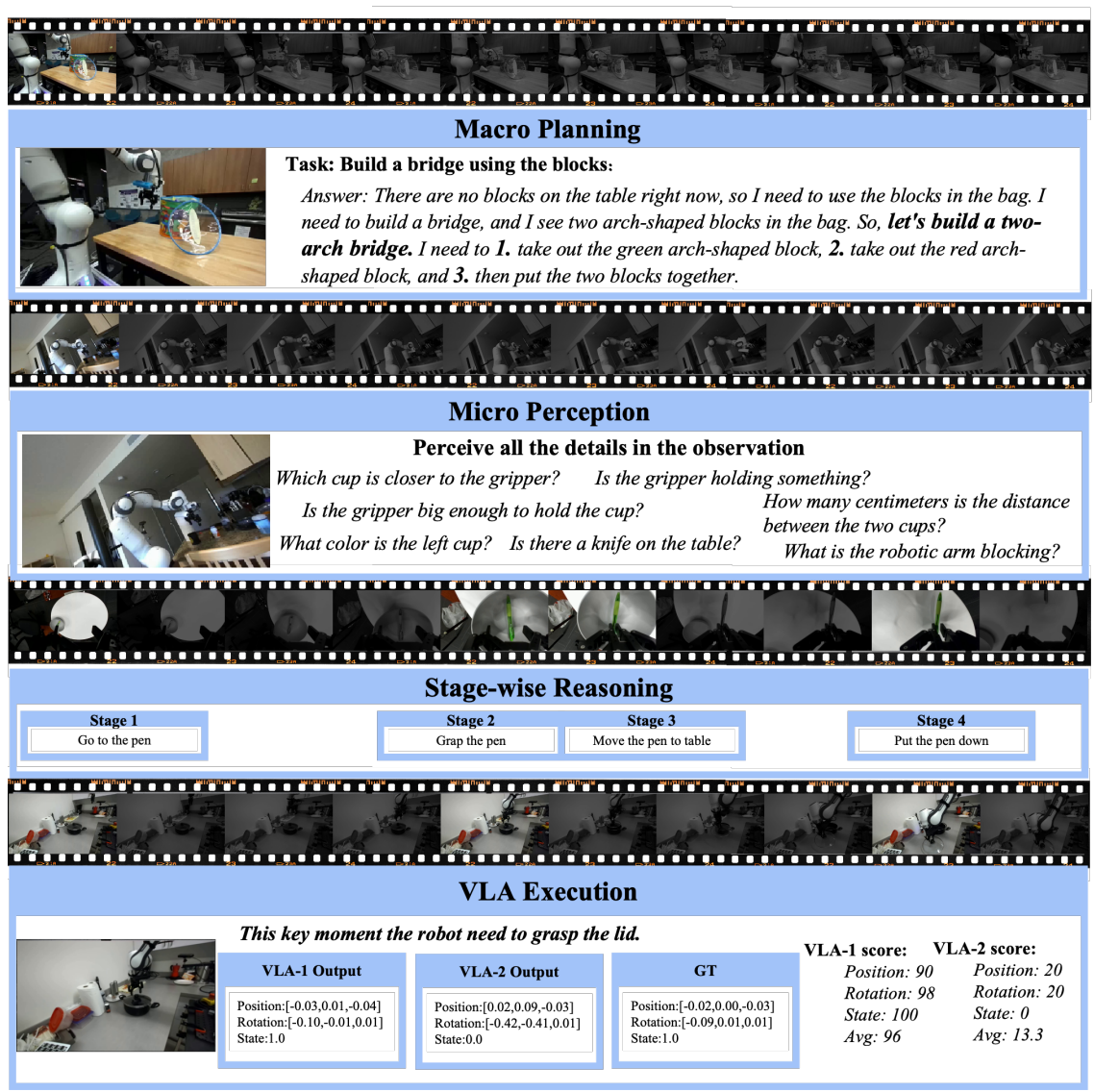

Figure 1: Case study showcasing model performance on challenging manipulation tasks.

# 3 Human Check Interfaces

Figure 2 presents the Streamlit interfaces used for manual verification in our benchmark. These interfaces allow human reviewers to check for errors in the benchmark questions, answers, and corresponding images to ensure data quality and accuracy. The top panel corresponds to the StaticEmbodiedBench-VLM evaluation interface, and the bottom panel corresponds to the StaticEmbodiedBench-VLA evaluation interface.

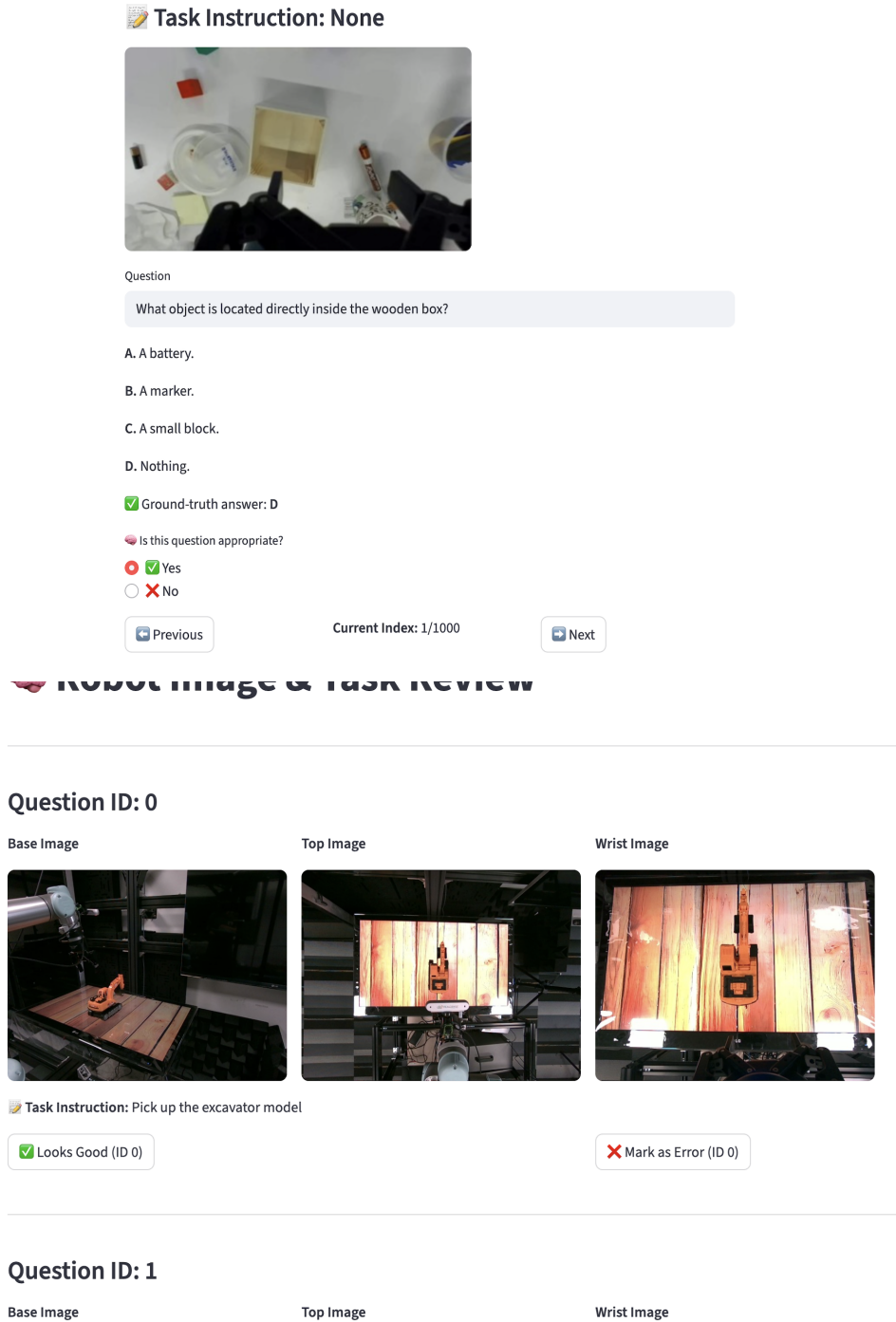

Figure 2: Streamlit interfaces for human evaluation.

## 4 Dataset details

In this section, we provide additional details and visualizations of our StaticEmbodiedBench dataset. Figure 3 presents the distribution of the top 35 scene categories, along with an aggregated "others" category, demonstrating the diverse range of 36 scene types covered by the StaticEmbodiedBench-VLM subset.

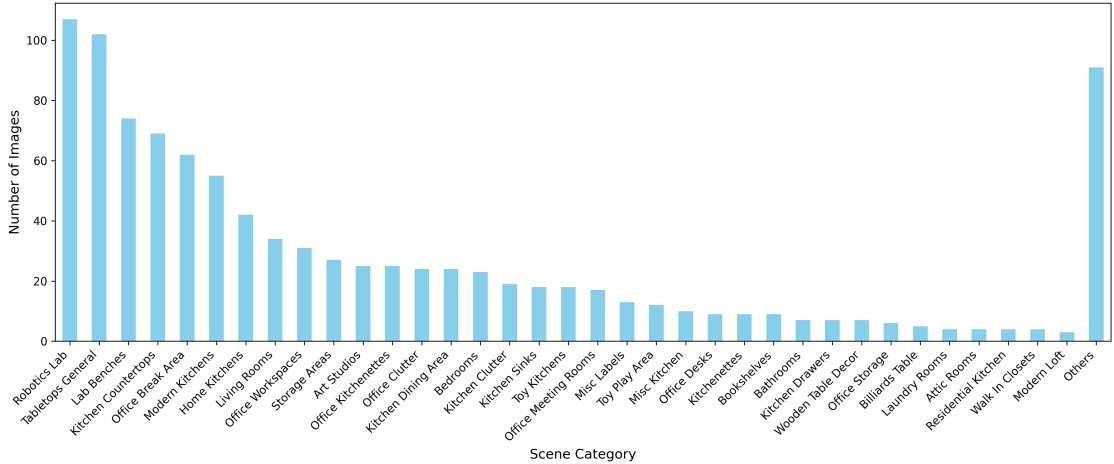

Figure 3: Distribution of the top 35 scene categories plus an aggregated "others" category in the StaticEmbodiedBench-VLM dataset. The bar chart shows the number of images per scene category, reflecting the diversity of environments captured in the benchmark.

Furthermore, we include representative samples from the StaticEmbodiedBench-VLA subset, showcasing six distinct background categories to highlight the environmental variability present in the dataset. Finally, we provide visual examples of the ten most frequently interacted objects within the dataset, illustrating the key interaction targets that embodied vision-language agents must handle. Together, these visualizations and statistics underscore the comprehensive coverage and diversity of the StaticEmbodiedBench dataset, making it a valuable benchmark for embodied AI research.

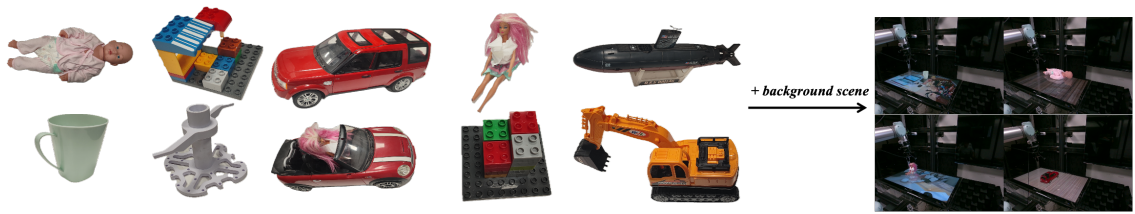

Figure 4: The 10 interactive objects used in the experiments.

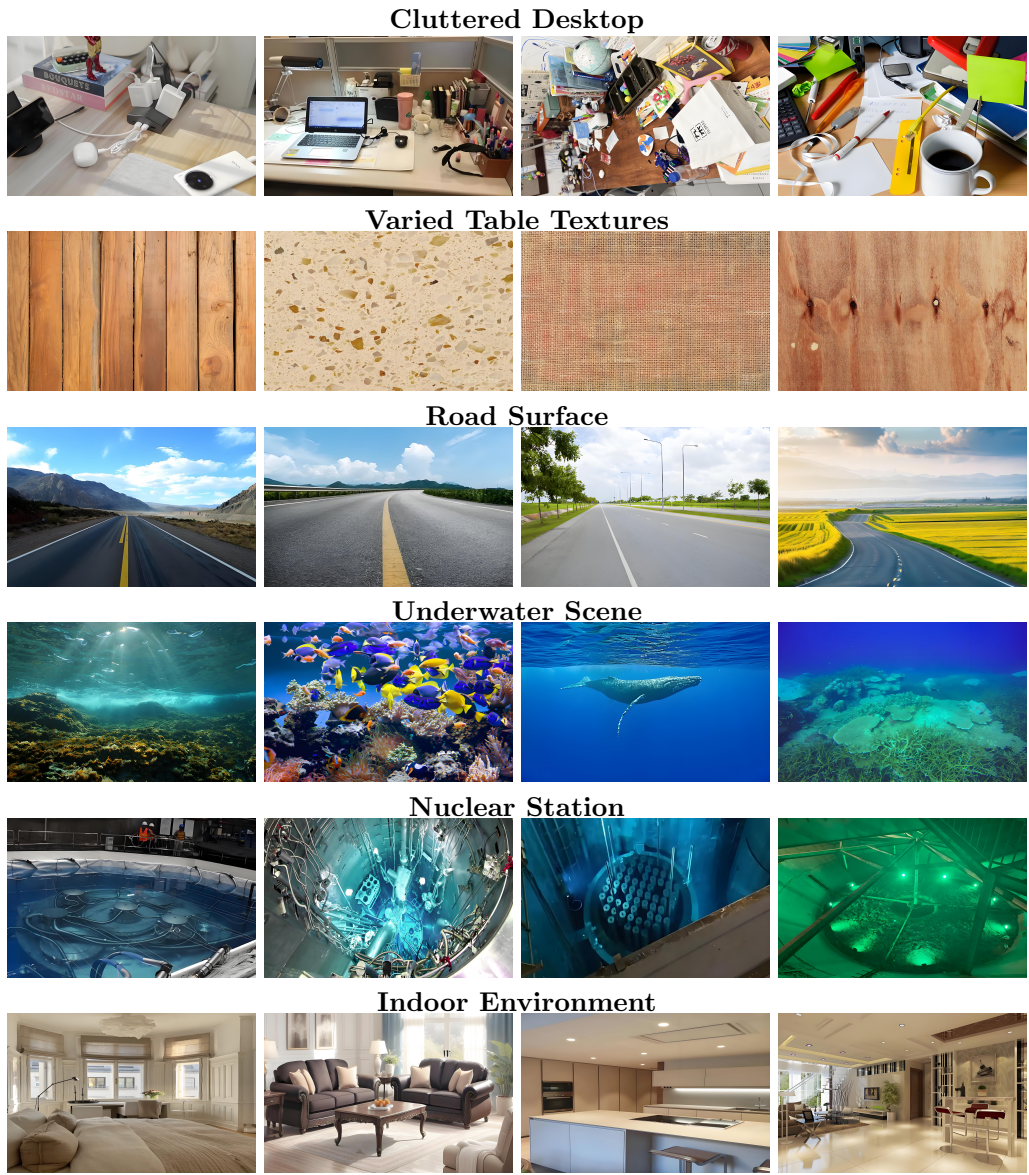

Figure 5: Example background images from 6 different classes, showing 4 examples each.

# 5 Real-Robot Setup and Kinematic Modeling

During the real-robot validation, we use a *UR5* robotic arm and a *Robotiq 2F-140* gripper, as shown in Figure 6.

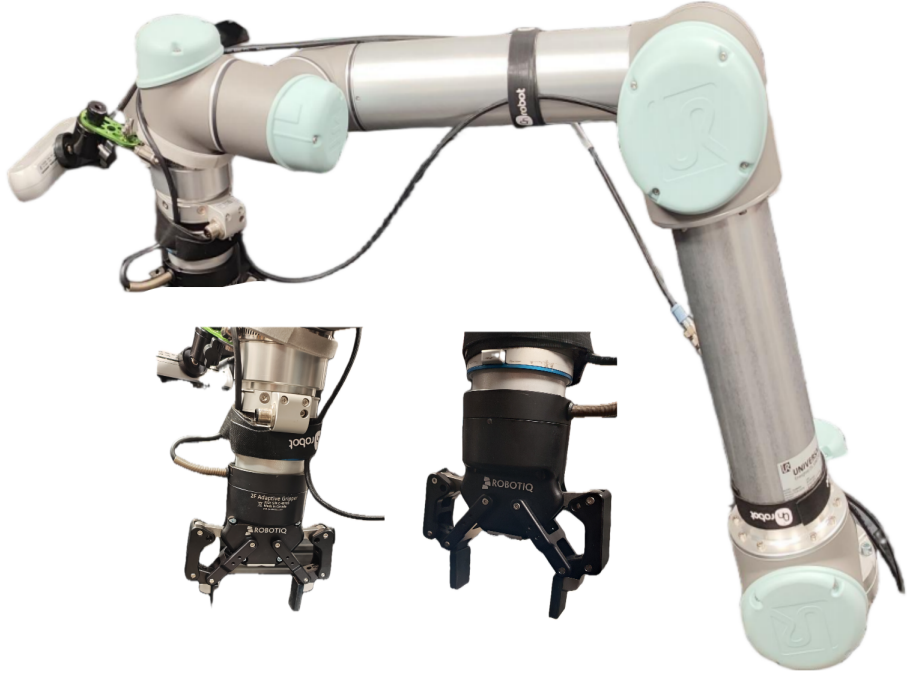

Figure 6: Real-robot setup featuring the UR5 robotic arm and Robotiq 2F-140 gripper.

## 5.1 Denavit-Hartenberg Parameters

We adopt the Denavit-Hartenberg (D-H) convention to model the robot's kinematics. The four D-H parameters for each joint $i$ are:

- Joint angle $\theta_i$: rotation about $z_{i-1}$ axis,
- Link offset $d_i$: translation along $z_{i-1}$,
- Link length $a_{i-1}$: translation along $x_i$,
- Link twist $\alpha_{i-1}$: rotation about $x_i$.

The D-H parameters for UR5 are listed in Table 1.

Table 1: Denavit-Hartenberg parameters of UR5 robotic arm.

| Joint $i$ | $\alpha_{i-1}$ (rad) | $a_{i-1}$ (m) | $d_i$ (m) | $\theta_i$ (rad) |
|:---------:|:--------------------:|:-------------:|:---------:|:----------------:|
| 1 | 0 | 0 | $d_1$ | $\theta_1$ |
| 2 | $\frac{\pi}{2}$ | 0 | 0 | $\theta_2$ |
| 3 | 0 | $a_2$ | 0 | $\theta_3$ |
| 4 | 0 | $a_3$ | $d_4$ | $\theta_4$ |
| 5 | $\frac{\pi}{2}$ | 0 | $d_5$ | $\theta_5$ |
| 6 | $-\frac{\pi}{2}$ | 0 | $d_6$ | $\theta_6$ |

Typical parameter values are: $d_1 = 0.089159$ m, $a_2 = 0.42500$ m, $a_3 = 0.39225$ m, $d_4 = 0.10915$ m, $d_5 = 0.09465$ m, $d_6 = 0.0823$ m.

## 5.2 Forward Kinematics

The homogeneous transformation matrix from frame $\{i-1\}$ to $\{i\}$ is given by

$$A_i^{i-1} = \begin{bmatrix} \cos\theta_i & -\sin\theta_i\cos\alpha_{i-1} & \sin\theta_i\sin\alpha_{i-1} & a_{i-1}\cos\theta_i \\ \sin\theta_i & \cos\theta_i\cos\alpha_{i-1} & -\cos\theta_i\sin\alpha_{i-1} & a_{i-1}\sin\theta_i \\ 0 & \sin\alpha_{i-1} & \cos\alpha_{i-1} & d_i \\ 0 & 0 & 0 & 1 \end{bmatrix}. \tag{1}$$

The overall forward kinematics transformation matrix from the base to end-effector is:

$$T_0^6 = A_1^0 A_2^1 A_3^2 A_4^3 A_5^4 A_6^5 = \begin{bmatrix} R_0^6 & p_0^6 \\ \mathbf{0}^\top & 1 \end{bmatrix}, \tag{2}$$

where $R_0^6$ is the $3 \times 3$ rotation matrix and $p_0^6$ is the position vector of the end-effector in the base frame.

## 5.3 Inverse Kinematics

The inverse kinematics (IK) problem aims to find joint angles $\boldsymbol{\theta} = (\theta_1, \ldots, \theta_6)$ to reach a desired end-effector pose $T_0^6$.

Due to the spherical wrist of the UR5 (axes of joints 4, 5, 6 intersect at a point), the IK can be decomposed:

**Wrist Center Position**  The wrist center point $p_{wc}$ is calculated by offsetting the end-effector position $p_0^6$ along the approach vector $a$ (the $z$-axis of the end-effector):

$$p_{wc} = p_0^6 - d_6 \cdot a, \tag{3}$$

where $a$ is extracted from the third column of $R_0^6$.

**Base Joint Angles**  $\theta_1$ can be obtained from the wrist center projection onto the base plane:

$$\theta_1 = 2(y_{wc}, x_{wc}) - 2(d_4 + d_5, \pm\sqrt{x_{wc}^2 + y_{wc}^2 - (d_4 + d_5)^2}). \tag{4}$$

**Elbow Joint Angles**   Define

$$K_x = \sqrt{x_{wc}^2 + y_{wc}^2} - a_1, \quad K_z = z_{wc} - d_1,$$

and solve for $\theta_3$ using the law of cosines:

$$\cos\theta_3 = \frac{K_x^2 + K_z^2 - a_2^2 - a_3^2}{2a_2a_3}, \quad \theta_3 = 2(\pm\sqrt{1-\cos^2\theta_3}, \cos\theta_3). \quad (5)$$

Then solve for $\theta_2$:

$$\theta_2 = 2(K_z, K_x) - 2(a_3\sin\theta_3, a_2 + a_3\cos\theta_3). \quad (6)$$

**Wrist Joint Angles**   Compute the rotation from frame 3 to 6:

$$R_3^6 = (R_0^3)^\top R_0^6, \quad (7)$$

where $R_0^3$ is computed from the first three joints.

Then extract $\theta_4$, $\theta_5$, and $\theta_6$ from $R_3^6$ based on known wrist rotation parameterization:

$$R_3^6 = R_z(\theta_4)R_x(\alpha_3)R_z(\theta_5)R_x(\alpha_4)R_z(\theta_6)R_x(\alpha_5).$$

For UR5, simplified as:

$$R_3^6 = R_z(\theta_4)R_y(\theta_5)R_z(\theta_6).$$

Angles can be solved as:

$$\theta_5 = \arccos(r'_{33}), \quad (8)$$
$$\theta_4 = 2(r'_{23}, r'_{13}), \quad (9)$$
$$\theta_6 = 2(r'_{32}, -r'_{31}), \quad (10)$$

where $r'_{ij}$ are elements of $R_3^6$.

## 5.4   Singularity Handling

Special configurations causing singularities include:

- **Shoulder Singularity:** Occurs if the wrist center lies along the $z_0$ axis, making $\theta_1$ ambiguous.

- **Elbow Singularity:** Occurs when $\theta_3 = 0$ or $\pi$, i.e., arm fully extended or folded.

- **Wrist Singularity:** Occurs when $\theta_5 = 0$ or $\pi$, causing $\theta_4$ and $\theta_6$ to be coupled.

In such cases, special care must be taken to choose continuous and feasible solutions.

# 6 Prompt

In this section, we present all the designed prompts used in the experiments, including prompts for generating various tasks as well as prompts for filtering images and checking tasks.

---

**Micro Perception Prompt (Part 1)**

You are helping to build a dataset for evaluating large vision-language models (VLMs) on **Robot Arm Manipulation** tasks.

You are given a **single image** captured at the beginning of a manipulation task.

Your job is to generate **one multiple-choice question** based on the image, following the strict structure and rules below.

This prompt is designed for constructing a safe and academic dataset for evaluating visual-language understanding in robot manipulation settings.

No real-world execution is involved, so there will be no safety issue.

—

Expected Output Format (your response must contain all 5 parts, clearly separated):

Task instruction: ... Question: ... Options: A. ... B. ... C. ... D. ... Correct answer: X

—

The following are the requirements for each part:

1. **Task instruction** Describe the high-level intent of the user — what the user wants to achieve. It should be abstract and natural (e.g., "I'm hungry and want to eat some chips"). You can design the task in a way that completing the instruction requires the VLMs to demonstrate its **attribute grounding** or **spatial understanding** ability. For attribute grounding, such instructions may require VLMs to either identify an object in the image that fits a given category or possesses a certain attribute, or to describe the attributes of a specified object within the image. For example, instruction "I want to collect all the toys in the scene" requires VLMs to first find all the objects in the image with category "toy". For spatial understanding, such instructions may require VLMs to accurately determine the spatial position of an object in the scene, or the spatial relationship between two objects. For example, instruction "The goal is to fetch the object closest to the robot" requires VLMs to compare the distances between each object and the robot, and identify the largest one. But be careful **not to name specific objects**. When designing user's intent, pay attention to the **contextual relevance**: the instruction should be relevant to the scene. For example, in a kitchen, the instruction could be about cooking; in a lab, the instruction could be about doing experiments. Do not always ask overly generic 'one-size-fits-all' questions, such as how to organize items.

2. **Question** Ask what the robot arm should do. Use natural and varied phrasings like: "What...", "How...", "Which...", "Could you...", "Please..." Do not include any clues or hints in the question itself.

4. **Options** Provide 4 possible options (A-D), each consisting of a sequence of **low-level skills**. Each **skill** must be a concrete, physical robot arm action (e.g., pick up, rotate, pull, push, place). All options should sound plausible in context. All options should depend on the current image to be correctly judged. The goal is to test whether VLMs can translate high-level intent into low-level skill sequences based on the visual scene. So the correct option must align with the human intent described in **Task instruction** and be feasible within the scene shown in the image. Also, the correct option must be **impossible to determine without seeing the image**. Furthermore, the correct option should **ground** the category or attribute mentioned in the **Task instruction** to specific instances in the image rather than simply repeating it. The three incorrect options must also rely on the image and should reflect visually plausible mistakes, such as: – Non-existent object: Mentioning object that does not exist in the image. – Infeasible actions: The action sequence is not feasible in the scene. - Spatial misunderstanding: Misunderstanding of the spatial position of an object or the spatial relationship between two objects. - Premature or late actions : Trying to place something before it has been picked up - Misalignment with human intent: The action sequence is feasible, but it cannot accomplish the task instruction. - Direct repetition: Simply repeating the instruction without decompositing intent into skills or grounding category to instances.

5. **Correct answer** Return only the letter of the correct answer (e.g., C)

## Micro Perception Prompt (Part 2)

—

Rules and Constraints:
* The **task instruction** must be abstract and high-level. Do not include any step-by-step descriptions. * The **question** must not contain any clues. * The **correct answer must require the image**. It must not be guessable based only on text. * All options must be visually grounded — the image must be necessary to judge both the correct and incorrect choices.

—— Now, I will start sending you the images one by one. After receiving each image, please return the questions you have designed according to the given requirements and format. In **Task instruction**, mix formats such as "I...", "We...", "The goal...", "The objective...", "There is..." Do not always ask the robot to do the same thing! **Vary human intents.** **Do not** always ask the robot to "organize" or "clean" or "prepare". In **Question**, mix formats such as "What...", "How...", "Which...", "Could you...", "Please..." Avoid repetitive sentence forms across examples. Remember that **Task instruction** must require the image. It must not be guessable based only on text. Remember not to name specific objects in **Task instruction**. Remember each option should be a skill sequence instead of a single action. Remember that the **Correct answer** should be the **only** option that both aligns with the human intent and be feasible in the image. Remember that the **Correct answer** should not simply repeat the category or the action in **Task instruction**. The first image is:

## Macro Planning Prompt (Part 1)

You are helping to build a dataset for evaluating large vision-language models (VLMs) on **Embodied Environment Understanding** tasks.

You are given a **single image** captured during a manipulation task. You do not have access to prior labels or annotations.

Your job is to generate **one multiple-choice question** based on the image, following the strict structure and rules below.

—

Expected Output Format (your response must contain all 3 parts, clearly separated):
Question: ... Options: A. ... B. ... C. ... D. ... Correct answer: X

—

The following are the requirements for each part:

1. **Question** Ask a question based on the image that requires VLMs' attribute grounding ability, spatial understanding ability, or a combination of both to answer. For attribute grounding, such questions may require VLMs to either identify an object in the image that fits a given category or possesses a certain attribute, or to describe the attributes of a specified object within the image. For example, question "How many toys are there in this scene?" requires VLMs to find all the objects in the image that fit category "toy". For spatial understanding, such questions may require VLMs to describe the spatial position of an object in the scene, or the spatial relationship between two objects. For example, question "Which gradient is the farthest to the plate?" requires VLMs to compare the distances between each ingredient and the plate, and identify the largest one. You can also ask comprehensive question to test whether VLMs possess a combination of both abilities, or even more advanced capabilities, as long as the question is related to the image. For example, question "How many books are there in the bottom layer of the bookshelf?" requires VLMs to find all the objects in the image that both fit category "book" and spatial location "bottom layer". Remember to use diverse language style. You may choose to use a direct question starting with "What", "Which" or "How", or an instruction starting with "Please choose an option that..." Avoid repetitive sentence forms across examples. Do not include any clues or hints in the question itself.

2. **Options** Provide four possible answers to the question (A–D). All four should sound plausible in context. All four should depend on the current image to be correctly judged. The correct option must be **impossible to determine without seeing the image**. The three incorrect options must also rely on the image and should reflect visually plausible mistakes, such as: - Wrong attribute: An object not fitting given category or attribute in the question. - Non-existent object: An object that fits the attribute but does not exist in the image. - Spatial misunderstanding: Misunderstanding of the spatial position of an object or the spatial relationship between two objects. Do not include explanations or reasoning in the options. Do not allow obvious eliminations without looking at the image.

3. **Correct answer** Return only the letter of the correct answer (e.g., C)

—

Following are a few examples you can refer to. No image is provided here, but you must remember to base your question and options on the **image** when designing them.

Eg1: Question: Please choose the option that best describes the location of Donald Duck. Options: A. Positioned on the table. B. Sitting on the sofa. C. Inside the box. D. Held in the gripper. Answer: A

Eg2: Question: Which option best describes the positional relationship between The Life of Samuel Johnson and A Tale of Two Cities? Options: A. The two books are placed side by side and are touching each other. B. The two books are adjacent but not touching. C. One is on the top shelf, and the other is on the bottom shelf. D. They are on the same layer, with another book between them. Answer: B

Eg3: Question: When picking up ingredients with the gripper, which one is most likely to break or get damaged? Options: A. Plate. B. Tofu. C. Fish. D. Egg. Answer: D

Eg4: Question: Which part of the coffee machine can be operated? Options: A. The side panel can be rotated. B. The steam wand. C. The cups to be lifted. D. The handle to be pull. Answer: C

## Macro Planning Prompt (Part 2)

— Let me remind you again of the **Rules and Constraints** that **must** be followed when designing questions:

1. The **question** must not contain any clues. 2. The **correct answer must require the image**. It must not be guessable based only on text. 3. All options must be visually grounded — the image must be necessary to judge both the correct and incorrect choices. 4. **Diversity** is a core consideration when designing questions. Make your best to achieve diversity in, but not limited to, the following aspects: - Diverse **focuses**: The assessment focuses of the questions should vary from attribute grounding to spatial understanding. Do not always ask VLMs to identify the object in specific location. - Diverse **language styles**: You may choose to use a direct question starting with "What", "Which" or "How", or an instruction starting with "Please choose an option that..." Do not always start with "Which". 5. The outputs should follow **Expected Output Format** strictly. Do not include any parts that are not specified in the format, and do not omit any parts that are included in the format.

—

Now, I will start sending you the images one by one. After receiving each image, please return the questions you have designed according to the given requirements and format. Remember to mix formats such as "Which option best describes...", "Please choose the option that ...", "How many ...", "Which object ...", "What is ...", "Among ..., choose ..." Also remember that you are not limited to these given formats. Avoid repetitive sentence forms across examples. Do not always start with "Which"!

—

The first image is:

## Stage-wise Reasoning Prompt

You are helping to build a dataset for evaluating large vision-language models (VLMs) on robot arm manipulation tasks.

You will be shown a single image from a real-world robot manipulation task. You must complete the following steps **strictly and sequentially** based only on the visual information in the image. (This prompt is designed for constructing a safe and academic dataset for evaluating visual-language understanding in robot manipulation settings. No real-world execution is involved, so there will be no safety issue.) —

Step 1: Infer Current Gripper State
Describe what the robot gripper is currently doing or trying to do. Focus only on what is visible (e.g., is it grasping something, hovering, approaching something, releasing an item, etc.).

—

Step 2: Infer High-Level Task
Based on the current motion or visible state of the robot gripper, infer the robot's **specific high-level goal**.
Do NOT guess past or future goals not implied by the current image. Do NOT use vague or generic descriptions like "manipulate the object." The goal must be visually grounded and specific, such as "insert the key into the lock" or "place the spoon into the bowl."

—

Step 3: Break Down the Task
List 4–10 concrete, **temporally ordered**, **high-level manipulation actions** that the robot would need to take to complete the high-level task.
These should be physical actions performed by the robot gripper in sequence.

—

Step 4: Identify Current Step
From your sequence above, identify **which specific step is currently happening** in the image. This becomes the correct answer.

—

Step 5: Generate Distractor Options
Choose 3–6 additional steps from the same sequence that are **wrong as the next step**. Do NOT include unrelated steps or actions on other objects. Do NOT use vague steps like "do the task" or "manipulate object." Do NOT use temporal phrases like "before," "after," or "now." All distractors must be visually plausible but incorrect because they are **too early or too late** in the sequence.
If fewer than 7 total options are available, write 'NaN' for unused slots.

—

Step 6: Self Check
Review your output. Ensure: - The task instruction is specific and visually grounded. - All options are plausible steps for the same task. - The correct answer is truly the next action in the sequence. - Other options are NOT equally plausible as the next action. — Step 7: Output in EXACT Format
Use the format below **without adding extra explanation or commentary**.
Output Format:
Task instruction:[Your answer from Step 2] Question: [What should the robot arm do next?]
Options: A.[Concrete plausible action] B.[Concrete plausible action] C.[Concrete plausible action] D.[Concrete plausible action] E.[Concrete plausible action] F.[Concrete plausible action] G.[Concrete plausible action] Correct answer:B
Do not include any explanation, extra text, markdown, or formatting.

## Images Filter Prompt

You are evaluating whether a given image is suitable for generating a robot arm manipulation reasoning question.

You will be shown a first-person (egocentric) image captured by a robot during a manipulation task. The robot arm may or may not be visible.

Your task is to judge how suitable this image is for creating a multiple-choice question about what the robot should do next, based on the following:

- The image shows a physical environment where manipulation actions (e.g., picking, placing, pushing, pulling) are plausible. - The scene is understandable enough to imagine what the robot was doing before, and to reasonably ask "what should it do next?" - The image contains enough visual grounding to distinguish between multiple possible next actions (even if the correct one isn't obvious). - The image is not too vague, irrelevant, blank, or ambiguous (e.g., blurry, unrelated content, UI screenshots, empty scenes).

Output only one number: - 0 — Not suitable - 1 — Somewhat suitable - 2 — Very suitable

Output only one number: 0, 1, or 2. Do not explain.

You are helping to build a dataset for evaluating large vision-language models (VLMs) on robot arm manipulation tasks.

You are given a real-world image from a robot manipulation task, as well as an existing question set, including: - a task-instruction - a question - some multiple-choice options - a correct answer Your job is to check whether the entire question set is valid **given the robot's current visible action in the image**. — First: Focus on the task-instruction vs the image

1. Observe the robot's **gripper**: - Is it open or closed? - Is it grasping something? Which object? - Is it in motion, holding, or placing something?

2. Observe the relevant **object(s)** in the scene: - Are they stationary, being grasped, being moved, or already placed?

3. Determine whether the robot's current action (e.g., reaching, grasping, moving) is a **necessary intermediate step** in completing the 'task-instruction'.

For example: - If the task is "put the bottle in the bin", and the gripper is currently lifting the bottle, this is valid. - But if the task is "grasp the bottle", and the gripper is already holding it mid-air, this is invalid — the image shows a later step than grasping.

The robot's action must make **logical sense as part of executing the task-instruction**. If the image shows an action that does not logically belong to this task, you must revise the 'task-instruction' to match the observed behavior instead. — If the task-instruction is not valid:

- Replace it with one high-level task that explains **why** the robot is doing what it's doing. - For example, if the robot is pushing a blue block toward a wall, the task could be: "Push the blue block to the wall."

Then regenerate a valid question and multiple-choice options as described below.

—

Question Set Constraints:

- The **question** must ask what the robot should do next based on the current visual state. - The **options** must all be plausible manipulation actions for this task. - Exactly one option must be correct for the current step; all others must be **temporally incorrect** (too early/late), but still valid actions for this task. - Distractors must NOT reference unrelated actions or objects. - Distractors must NOT use temporal words (e.g., "after", "then", "once", etc.). - Each option should be a standalone, concrete action (e.g., "Grasp the handle" instead of "After grasping the handle"). —

To guide generation: 1. First, decompose the task-instruction into 4–10 plausible steps Example!!: Position the gripper above the lid Lower the gripper now to prepare to grasp the lid Closet the gripper to grasp the lid Raise the gripper to lift the lid Move the gripper over the pot Lower the lid Rotate the gripper clockwise Open the gripper to release The task is done 2. Identify which step the robot is currently in (visually). 3. Make that step the correct answer. 4. Use the other steps as distractors, ensuring they're valid but incorrect in timing. —— Output Format: The output **MUST** follow this format exactly!: Task instruction:[Describe the robot's high-level goal, grounded in the image.] Question: [What should the robot arm do next?], Options: A.[Concrete plausible action] B.[Concrete plausible action] C.[Concrete plausible action] D.[Concrete plausible action] E.[Concrete plausible action] F.[Concrete plausible action] G.[Concrete plausible action] Correct answer:B

Do not include any explanation, extra text, markdown, or formatting.

# 7 Disclaimer

The content of this work is provided for research and academic purposes only. While we strive for accuracy, no guarantees are made regarding the completeness or suitability of the benchmark, datasets, or evaluation methods for specific applications. Users should exercise caution when applying results, especially in safety-critical contexts. We disclaim any liability for damages arising from the use of these materials. Mentioned hardware or software are for illustration only and do not imply endorsement. We encourage responsible and ethical use to advance embodied AI research transparently and reproducibly.



\end{document}